# Nonnegative Matrix Factorization with Toeplitz Penalty

Matthew A. Corsetti[1,*] and Ernest Fokoué[2]

[1] *Department of Biostatistics and Computational Biology, University of Rochester, Rochester, NY 14627, United States*
[2] *School of Mathematical Sciences, Rochester Institute of Technology, Rochester, NY 14623, United States*
*\*Corresponding author*: corsetti.matthew@gmail.com

**Abstract.** *Nonnegative Matrix Factorization* (NMF) is an unsupervised learning algorithm that produces a linear, parts-based approximation of a data matrix. NMF constructs a nonnegative low rank basis matrix and a nonnegative low rank matrix of weights which, when multiplied together, approximate the data matrix of interest using some cost function. The NMF algorithm can be modified to include auxiliary constraints which impose task-specific penalties or restrictions on the cost function of the matrix factorization. In this paper we propose a new NMF algorithm that makes use of non-data-dependent auxiliary constraints which incorporate a Toeplitz matrix into the multiplicative updating of the basis and weight matrices. We compare the facial recognition performance of our new *Toeplitz Nonnegative Matrix Factorization* (TNMF) algorithm to the performance of the *Zellner Nonnegative Matrix Factorization* (ZNMF) algorithm which makes use of data-dependent auxiliary constraints. We also compare the facial recognition performance of the two aforementioned algorithms with the performance of several preexisting constrained NMF algorithms that have non-data-dependent penalties. The facial recognition performances are evaluated using the Cambridge ORL Database of Faces and the Yale Database of Faces.

**Keywords.** Nonnegative matrix factorization; Auxiliary constraints; Toeplitz matrix; Zellner *g*-Prior; Image processing; Facial recognition; Subspace methods

**MSC.** 62H35

**Received:** March 16, 2018    **Accepted:** June 7, 2018

## 1. Introduction

In the last several decades, facial detection and recognition via algorithmic-based approaches have become quite popular. These tasks are often complicated due to variation in illumination, orientation, emotional expression and physical location of a face within an image. With the rise of big data, facial databases have become massively large. This has further complicated the already complex tasks of facial detection and recognition. The need to process these large and complex datasets has caused a surge in the utilization of subspace methods, such as nonnegative matrix factorization (Paatero and Tapper [12], and Seung and Lee [16]). These methods are used for the purpose of reducing data dimensionality while simultaneously identifying and preserving latent underlying structure in the dataset.

For facial databases, the data matrix $\mathbf{X}$ contains $p$ vectorized images, each containing $n$ pixel value intensities. The NMF algorithm produces an approximate representation of the nonnegative data matrix $\mathbf{X} \in \mathbb{R}^{n \times p}$ in the form of a matrix product:

$$\mathbf{WH} \approx \mathbf{X}. \tag{1}$$

The NMF problem can be generalized as follows. Given a nonnegative data matrix $\mathbf{X} \in \mathbb{R}^{n \times p}$ and a positive integer $k \ll \min(n, p)$, find nonnegative matrices $\mathbf{W} \in \mathbb{R}^{n \times k}$ and $\mathbf{H} \in \mathbb{R}^{k \times p}$ to minimize a cost function in the form of some distance measure such as:

$$f(\mathbf{W}, \mathbf{H}) = \frac{1}{2} \|\mathbf{X} - \mathbf{WH}\|_F^2. \tag{2}$$

The matrix product consists of a nonnegative basis matrix $\mathbf{W} \in \mathbb{R}^{n \times k}$ and a nonnegative matrix of weights or coefficients $\mathbf{H} \in \mathbb{R}^{k \times p}$. The basis matrix $\mathbf{W}$ contains $k$ basis column vectors, each of length $n$. These basis vectors or basis images contain the underlying features of the dataset $\mathbf{X}$. Additive linear combinations of the basis images are obtained when taking the product of $\mathbf{W}$ and $\mathbf{H}$ because of the nonnegativity constraints. Each of the linear combinations represents or approximates a face in the data matrix $\mathbf{X}$. It is intuitive to associate the process with combining individual parts to form a whole face.

The nonnegative matrix factorization algorithm was first proposed by Paatero and Tapper in 1994 under the name "Positive Matrix Factorization" (Paatero and Tapper [12]). Lee and Seung proposed an efficient multiplicative updating algorithm (Seung and Lee [16]) as well as using the Kullback-Leibler divergence as an alternative to the Euclidean distance for the cost function (Lee and Seung [10]). There are many published works which explore various cost functions on the matrix factorization (Cichocki *et al*. [3], Wang *et al*. [19], Hamza and Brady [7], and Sr and Dhillon [17], Xue *et al*. [22], Sandler and Lindenbaum [15]). Many different minimization methods for the solution of (2) have been explored for the purpose of decreasing the convergence time of Lee and Seung's iterative algorithm (Lin [11], Gonzalez and Zhang [6], Zdunek and Cichocki [23]). $\mathbf{W}$ and $\mathbf{H}$ are initialized with nonnegative random values in Lee and Seung's multiplicative iterative algorithm (Seung and Lee [16]). Various alternative initialization strategies for $\mathbf{W}$ and $\mathbf{H}$ have been proposed for the purpose of increasing the speed of convergence or to manipulate convergence to a specific desired result (Wild [20], Wild et al. [21], and Boutsidis and Gallopoulos [1]).

Various authors have adapted the NMF algorithm by applying auxiliary constraints to the matrix **W** and/or **H**. Often these constraints take the form of smoothness constraints (Piper et al. [14], Pauca et al. [13], and Chen and Cichocki [2]) or sparsity constraints (Hoyer [8], and Hoyer [9]). Recently a set of data-dependent auxiliary constraints was explored (Corsetti and Fokoué [4]). These types of auxiliary constraints are usually incorporated so as to take into account prior information about the application under examination or to guarantee desired characteristics in the solution for the **W** and **H** matrices. The secondary constraints are enforced through the use of penalty terms which extend the cost function of equation (2) as follows:

$$f(\mathbf{W},\mathbf{H}) = \frac{1}{2}\|\mathbf{X} - \mathbf{W}\mathbf{H}\|_F^2 + \alpha J_1(\mathbf{W}) + \beta J_2(\mathbf{H}). \qquad (3)$$

For equation (3), $J_1(\mathbf{W})$ and $J_1(\mathbf{W})$ are the penalty terms and $\alpha$ and $\beta$ ($0 \leq \alpha \leq 1$ and $0 \leq \beta \leq 1$) are the regularization parameters which serve to control the trade-off between the approximation error of **WH** in approximating **X** and the auxiliary constraints. This method is commonly referred to as "Constrained Nonnegative Matrix Factorization".

We propose a novel constrained NMF algorithm by incorporating a Toeplitz matrix in the penalty terms of equation (3). We refer to this algorithm as "Toeplitz Nonnegative Matrix Factorization" (TNMF). We compare the recognition performance of the TNMF algorithm with the recognition performance of the Zellner Nonnegative Matrix Factorization (ZNMF) algorithm (Corsetti and Fokoué [4]) which makes use of data-dependent penalties, unlike the TNMF algorithm. We also compare the recognition performance of the TNMF algorithm with several other constrained NMF algorithms (Pauca *et al*. [13], and Wang et al. [19]). We assess the recognition performance of the aforementioned algorithms using the Cambridge ORL Database of Faces as well as the Yale Database of Faces. We find that the ZNMF algorithm outperforms the constrained NMF algorithms for the ORL dataset simulations and the TNMF algorithm outperforms the other NMF algorithms for the Yale dataset simulations.

## 2. Constrained Nonnegative Matrix Factorization

Constrained Nonnegative Matrix Factorization extends the cost function of traditional NMF shown in (2) to that of (3). The functions $J_1(\mathbf{W})$ and $J_2(\mathbf{H})$ represent penalty terms that are meant to constrain the solution of (3). $\alpha$ and $\beta$ are the respective regularization parameters of $J_1(\mathbf{W})$ and $J_2(\mathbf{H})$ such that $0 \leq \alpha \leq 1$ and $0 \leq \beta \leq 1$. In accordance with (Pauca *et al*. [13]) we set:

$$J_1(\mathbf{W}) = \|\mathbf{W}\|_F^2 \qquad (4)$$

and

$$J_2(\mathbf{H}) = \|\mathbf{H}\|_F^2. \qquad (5)$$

We enforce smoothness in the basis matrix W and statistical sparseness (Hoyer [8]) in the weights matrix **H** by setting $J_1(\mathbf{W})$ and $J_2(\mathbf{H})$ equal to the square of their respective Frobenius norms (Pauca *et al*. [13]). Using the penalties shown in (4) and (5), the standard NMF multiplicative updating equations for **W** and **H** (Seung and Lee [16]) are modified as follows:

$$\mathbf{W}^{(t+1)} = \mathbf{W}^{(t)} \frac{\mathbf{X}(\mathbf{H}^{(t)})^{\mathrm{T}}}{\mathbf{W}^{(t)}\mathbf{H}^{(t)}(\mathbf{H}^{(t)})^{\mathrm{T}} + \alpha \frac{\partial}{\partial \mathbf{W}} J_1(\mathbf{W}^{(t)})} \quad (6)$$

and

$$\mathbf{H}^{(t+1)} = \mathbf{H}^{(t)} \frac{(\mathbf{W}^{(t+1)})^{\mathrm{T}}\mathbf{X}}{(\mathbf{W}^{(t+1)})^{\mathrm{T}}\mathbf{W}^{(t+1)}\mathbf{H}^{(t)} + \beta \frac{\partial}{\partial \mathbf{H}} J_2(\mathbf{H}^{(t)})} \quad (7)$$

where $t$ denotes the iteration.

## 3. Zellner Nonnegative Matrix Factorization

The inspiration for the ZNMF algorithm and its details are described in (Corsetti and Fokoué [4]). The ZNMF algorithm incorporates Zellner's g-prior (Zellner [24]) and alters the objective function of equation 3 by setting:

$$J_1(\mathbf{W}) = \frac{1}{g}\mathrm{trace}(\mathbf{W}^{\mathrm{T}}\mathbf{X}\mathbf{X}^{\mathrm{T}}\mathbf{W}) \quad (8)$$

and

$$J_2(\mathbf{H}) = \frac{1}{g}\mathrm{trace}(\mathbf{H}\mathbf{X}^{\mathrm{T}}\mathbf{X}\mathbf{H}^{\mathrm{T}}) \quad (9)$$

where

$$g = \max(n, p^2). \quad (10)$$

Consequently,

$$\frac{\partial}{\partial \mathbf{W}} J_1(\mathbf{W}) = 2\mathbf{X}\mathbf{X}^{\mathrm{T}}\mathbf{W} \quad (11)$$

and

$$\frac{\partial}{\partial \mathbf{H}} J_2(\mathbf{H}) = 2\mathbf{H}\mathbf{X}^{\mathrm{T}}\mathbf{X}. \quad (12)$$

For the ZNMF algorithm, the updating equations for **W** and **H** are:

$$\mathbf{W}^{(t+1)} = \mathbf{W}^{(t)} \frac{\mathbf{X}(\mathbf{H}^{(t)})^{\mathrm{T}}}{\mathbf{W}^{(t)}\mathbf{H}^{(t)}(\mathbf{H}^{(t)})^{\mathrm{T}} + (\frac{\alpha}{g})\mathbf{X}\mathbf{X}^{\mathrm{T}}(\mathbf{W})^{(t)}} \quad (13)$$

and

$$\mathbf{H}^{(t+1)} = g\mathbf{H}^{(t)}\left[\frac{(\mathbf{W}^{(t+1)})^{\mathrm{T}}\mathbf{X}}{g(\mathbf{W}^{(t+1)})^{\mathrm{T}}\mathbf{W}^{(t+1)}\mathbf{H}^{(t)} + \beta\mathbf{H}^{(t)}\mathbf{X}^{\mathrm{T}}\mathbf{X}}\right]. \quad (14)$$

## 4. Toeplitz Nonnegative Matrix Factorization

Tipping [18] demonstrated the possibility of deriving sparse representations in kernel regression via suitably specified Gamma hyperpriors on the independent precisions of a Gaussian prior on the weights of the kernel expansion. Fokoué [5] explored a modified version of Tipping [18] by using structured matrix of the form

$$\begin{bmatrix} 1 & \rho & \cdots & \rho \\ \rho & 1 & \rho & \vdots \\ \vdots & \rho & \ddots & \rho \\ \rho & \cdots & \rho & 1 \end{bmatrix}. \tag{15}$$

In this paper, we adopt a different approach that seeks to achieve sparsity by directly exploiting the properties of Toeplitz type structured matrices. We argue that our approach makes sense for both variable selection in linear models and atom selection in kernel methods, because all our proposed matrices are reminiscent of matrices encountered in the analysis of autoregressive time series models. Indeed, such matrices do inherently capture correlations of varying degrees and can therefore isolate variables or atoms that are more dominant, leaving the least dominant ones as secondary members of a group. In a sense, our selected features are somewhat like representatives of a group with which the other members have a strong correlation.

$$s_{ij} = \begin{cases} 1 & \text{if } i = j \\ \rho^{|i-j|} & \text{if } i \neq j. \end{cases}$$

We specify a Gaussian prior for **w** with a variance-matrix that is structured *à* la Toeplitz, namely

$$\mathbf{w} \sim \mathcal{N}_n(\mathbf{0}, r\mathbf{S})$$

with

$$\Sigma_n \equiv \begin{bmatrix} 1 & \rho & \rho^2 & \cdots & \rho^{n-2} & \rho^{n-1} \\ \rho & 1 & \rho & \rho^2 & \cdots & \rho^{n-2} \\ \rho^2 & \rho & 1 & \ddots & \ddots & \vdots \\ \vdots & \rho^2 & \ddots & 1 & \ddots & \rho^2 \\ \rho^{n-2} & \vdots & \ddots & \ddots & 1 & \rho \\ \rho^{n-1} & \rho^{n-2} & \cdots & \rho^2 & \rho & 1 \end{bmatrix} \tag{16}$$

where $c$ is constant.

This can be substantially improved upon by using

$$s_{ij} = \begin{cases} 1 & \text{if } i = j \\ (-1)^{|i-j|+1} \left(\frac{\rho}{|i-j|}\right)^{v|i-j|} & \text{if } i \neq j \end{cases}$$

This new variance-matrix structure has the inherent property of creating a banded matrix so that associations are gradually lessened.

## 5. Experimental Results

Simulations were carried out to compare the facial recognition performances of the CNMF, TNMF and ZNMF algorithms against the facial recognition performances of traditional NMF, Local NMF (LNMF), Fisher NMF (FNMF), Principle Component Analysis and Principle

Component Analysis NMF (PNMF) using the results of Wang et al. [19] which carried out simulations on the ORL dataset exclusively. We further compared the facial recognition capabilities of the CNMF, TNMF and ZNMF algorithms using the Yale dataset.

The Cambridge ORL Database contains a total of 400 grey-scale images. 10 pictures were taken of each of the forty subjects. 36 of the 40 subjects are male and 4 are female. Each image contains a single face with varying levels of illumination, degree of rotation and emotional expression. In an effort to improve computational efficiency, the resolutions of the ORL images were reduced from $112 \times 92$ to $28 \times 23$ in accordance with Wang et al. [19], which found that reducing the image resolution to 25% the original size did not have a substantial impact on the recognition performance of the algorithms. For each of the ORL simulations, the training dataset $\mathbf{X}_{\geq 0}^{644 \times 200}$ consisted of 5 randomly selected images from each of the forty subjects making for a training set of 200 images each with 644 pixel value intensities. The test set for each simulation contained the remaining 200 unselected images.

The Yale Database consists of 165 grey-scale images. 11 pictures were taken of each of the 15 subjects. 14 of these subjects are male while only 1 is female. These images have varying levels of illumination and emotional expression; however, unlike the ORL faces, they are all forward-facing. A rather apparent difference between the two databases is the severity of the contrast between the background of the images and the boarder of the faces. This contrast is severe in the Yale Database and substantially less so in the ORL Database. The resolutions of the Yale images were also reduced from $243 \times 320$ to $61 \times 80$ in an attempt to improve the computational efficiency. In each of the Yale simulations, the training dataset $\mathbf{X}_{\geq 0}^{4880 \times 90}$ contained 6 randomly selected images from each of the 15 subjects, resulting in a training set of 90 images with 4,880 pixel intensities each. The subsequent test sets for each of the Yale simulations were composed of the remaining 75 images. Figures 1 and 2 depict the reduction in resolution for a subset of faces for each of the two datasets.

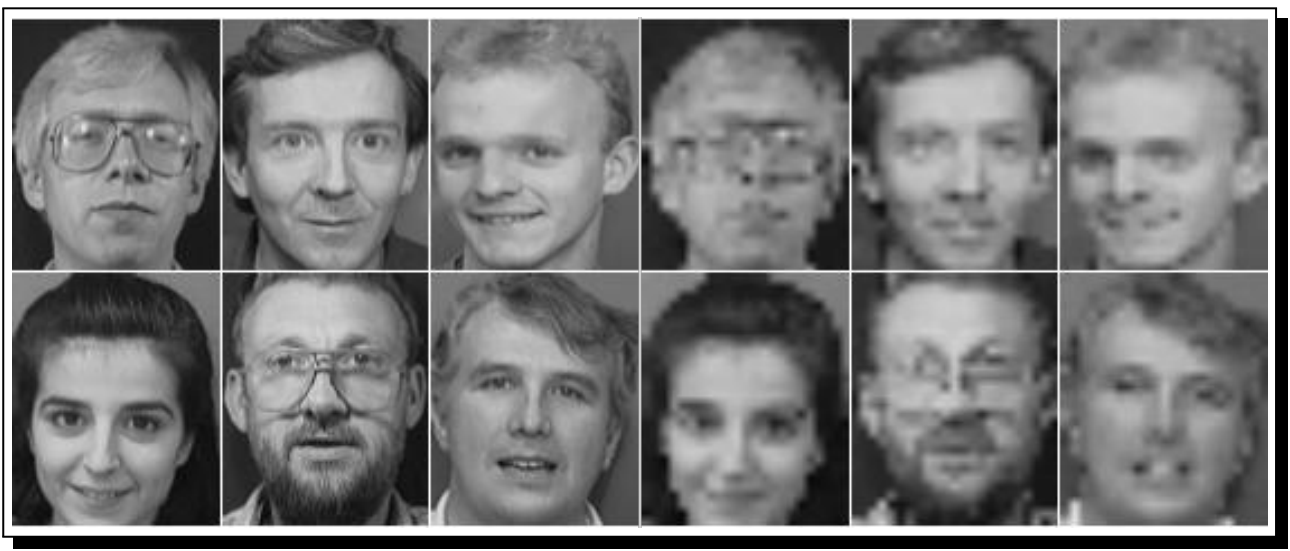

**Figure 1.** (left) 6 ORL faces at full $112 \times 92$ resolution; (right) 6 ORL faces at reduced $28 \times 23$ resolution

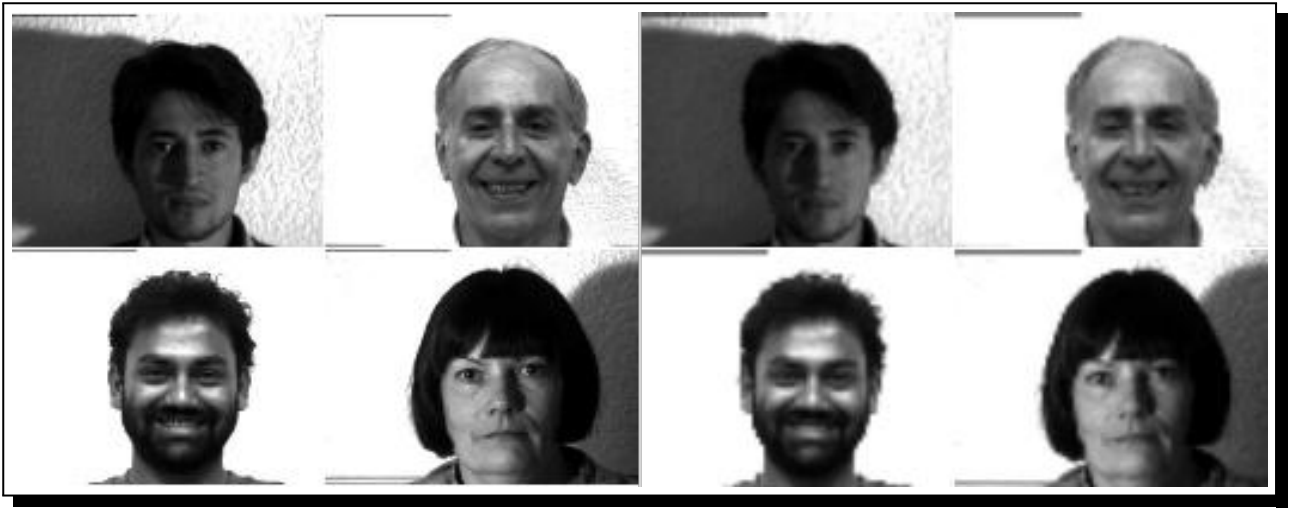

**Figure 2.** (left) 4 Yale faces at full 243 × 320 resolution; (right) 4 Yale faces at reduced 61 × 80 resolution

The effects on the facial recognition performance of the $\rho$ parameter and the $g$-Prior parameter in the TNMF and ZNMF algorithms respectively were investigated in great detail via extensive computer simulation, as were the $\alpha$ and $\beta$ parameters. The $\alpha$ and $\beta$ parameters were constrained via the following relationship:

$$\alpha = 1 - \beta \qquad (17)$$

such that $0 \leq \alpha \leq 1$ and $0 \leq \beta \leq 1$.

For each algorithm and in each dataset, 25 replications were carried out at each of the unique parameter setting combinations. In each of these replications a new training set, consisting of half of the images, was randomly sampled and used in calculating the average recognition performances of the CNMF, ZNMF and TNMF algorithms. The same seven factorization ranks, $k \in \{16, 25, 36, 49, 64, 81, 100\}$, were considered for each algorithm and dataset.

The optimal $\alpha$ and $\beta$ parameter settings for the CNMF algorithm were obtained by searching across a space from 0 to 1 in increments of 0.01. The results were smoothed using a loess curve with a 0.50 degree of smoothing parameter. The smoothed results are displayed in Figure 3. The large point along each of the factorization rank lines denotes the point at which the maximum recognition performance occurs for that specific factorization rank.

The initial search space for the optimal $\alpha$, $\beta$ and $g$-prior parameter settings of the ZNMF algorithm, using the ORL dataset was quite vast and not particularly granular. In this initial search space, a factorization rank of $k = 16$ was selected. The g-prior parameter was searched across an interval from 100 to 10,000 in increments of 100; while the $\alpha$ parameter was explored across an interval from 0 to 1 in increments of 0.01. As stated previously, 25 replications were carried out at each of these unique parameter settings. The resulting surface is displayed in the top left of Figure 4. A promising region, defined by a $g$-prior less than 100 and $\alpha$ value between 0.20 and 0.80 is quite noticeable in the initial explorative surface. As a result, more granular parameter searches were carried out across the seven factorization ranks using values for the *g*-prior from 75 to 90 in increments of 0.50 and values of 0.20 to 0.80 in increments of 0.01 for $\alpha$. The most promising surface was generated using $k = 36$ and is displayed in the top right of Figure 4. The optimal *g*-prior, $\alpha$ and $\beta$ settings were 83, 0.44 and 0.56 respectively.

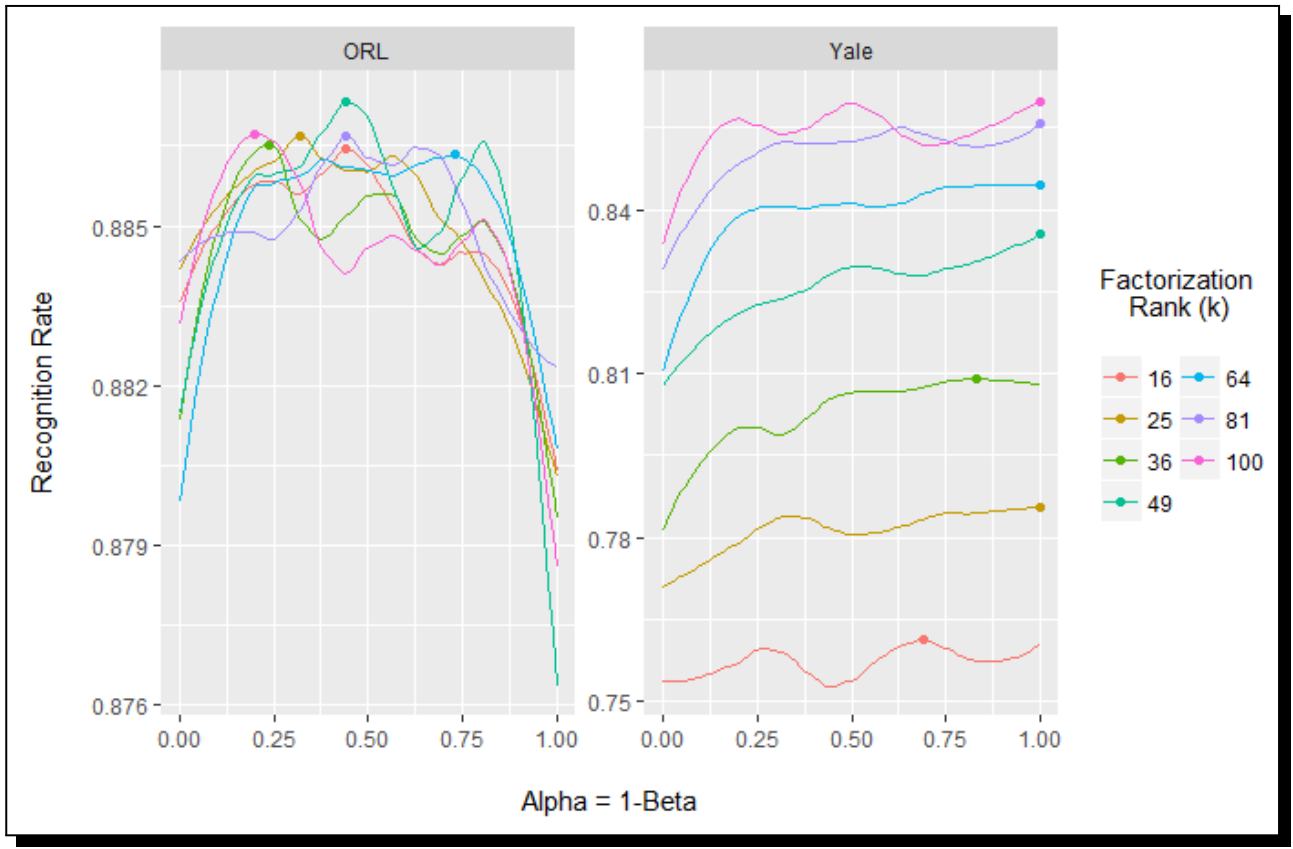

**Figure 3.** Recognition Performance of CNMF algorithm across various $\alpha$ and $\beta$ parameter values for the ORL Dataset (left) and Yale Dataset (right)

A similar approach was used in finding the optimal settings for the $\alpha$, $\beta$ and *g*-prior parameters for the Yale dataset. The $\alpha$ parameter was explored over a space from 0 to 1 in increments of 0.02 while the *g*-prior parameter was explored over a space from 200 to 10,000 in increments of 50. Again, an initial, broad explorative surface was generated using a factorization rank of $k = 16$ and is provided in the lower left of Figure 4. A promising area of the initial broad surface is defined by a *g*-prior value between 3,500 and 4,500 and an $\alpha$ value between 0.01 and 0.30. Simulations were carried out within the aforementioned parameter subspace across the seven different factorization ranks. $k = 100$ was found to be the optimal setting for the factorization rank. The resulting surface is displayed in the bottom right of Figure 4. The optimal *g*-prior, $\alpha$ and $\beta$ settings were 3500, 0.00 and 1.00, respectively.

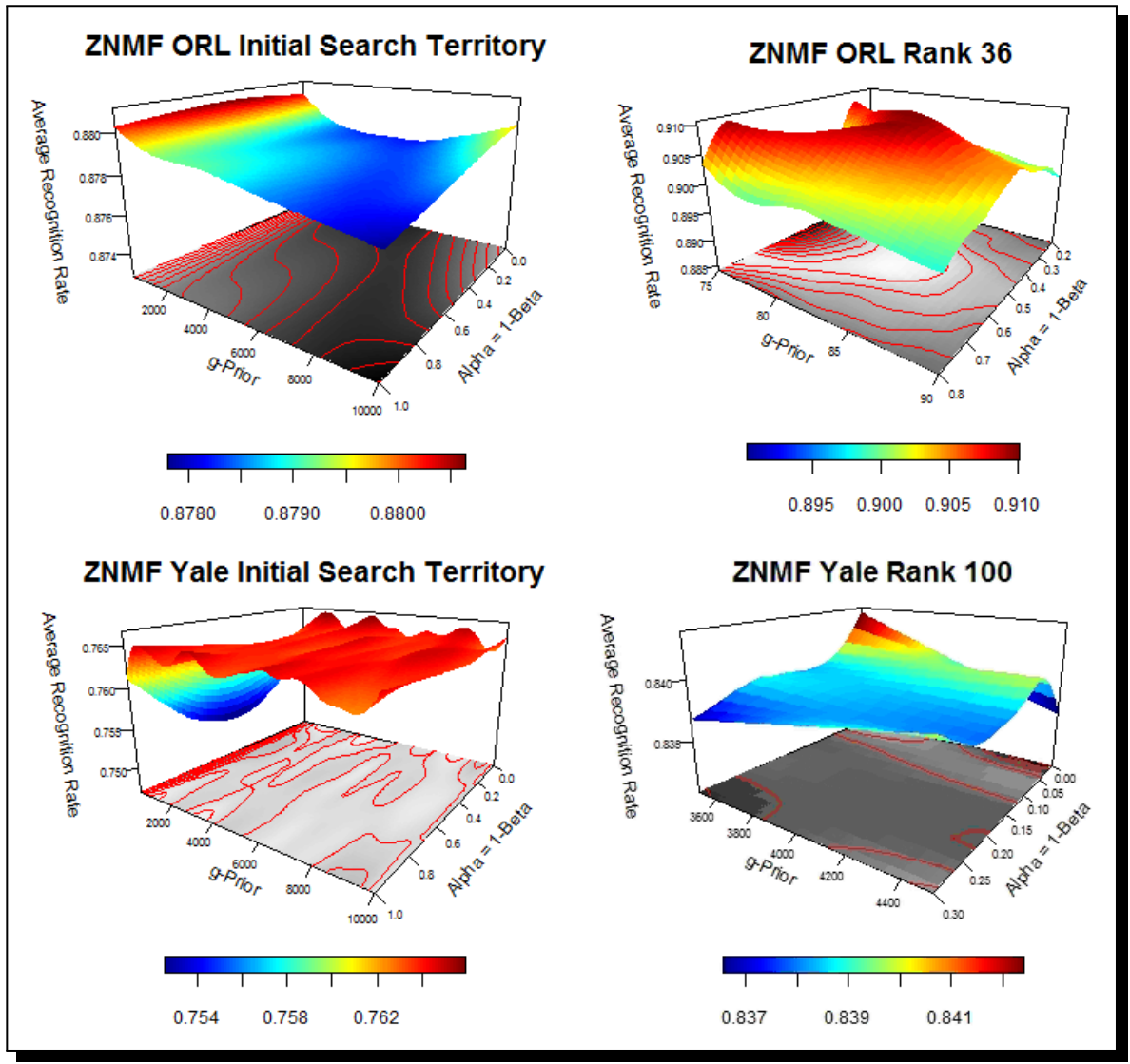

**Figure 4.** (top left) The initial search region for the parameters of the ZNMF algorithm using the ORL dataset with $k = 16$; (top right) The most promising granular search surface using the ZNMF algorithm with $k = 36$ for the ORL dataset; (bottom left) The initial search region for the parameters of the ZNMF algorithm using the Yale dataset with $k = 16$

The $\rho$ tuning parameter of the TNMF algorithm, as stated previously, was constrained to be $0 \leq \rho \leq 1$; and as a result, granular performance surfaces were constructed in relatively little time for the ORL dataset. The $\alpha$ and $\rho$ parameters were both explored over an interval from 0 to 1 in increments of 0.02 across all seven factorization ranks. The optimal surface was generated using a factorization rank of 25 and is displayed in the top of Figure 5. Optimal $\rho$, $\alpha$ and $\beta$ settings were found to be 0.06, 0.68 and 0.32 respectively.

An initial search surface was generated for the Yale dataset using the TNMF algorithm. The $\alpha$ and $\rho$ parameters were both searched over an interval from 0 to 1 in increments of 0.02. This surface is displayed in the bottom left of Figure 5. Again, 25 replications were carried out at each of the unique parameter settings so as to acquire an average predictive performance. A more detailed surface was produced by searching over $\alpha$ values from 0.30 to 0.70 in increments of 0.02 and $\rho$ values from 0.01 to 0.20 in increments of 0.01. The most promising factorization rank was discovered to be $k = 81$ using the TNMF algorithm on the Yale dataset. The resulting surface is provided in the lower left of Figure 5.

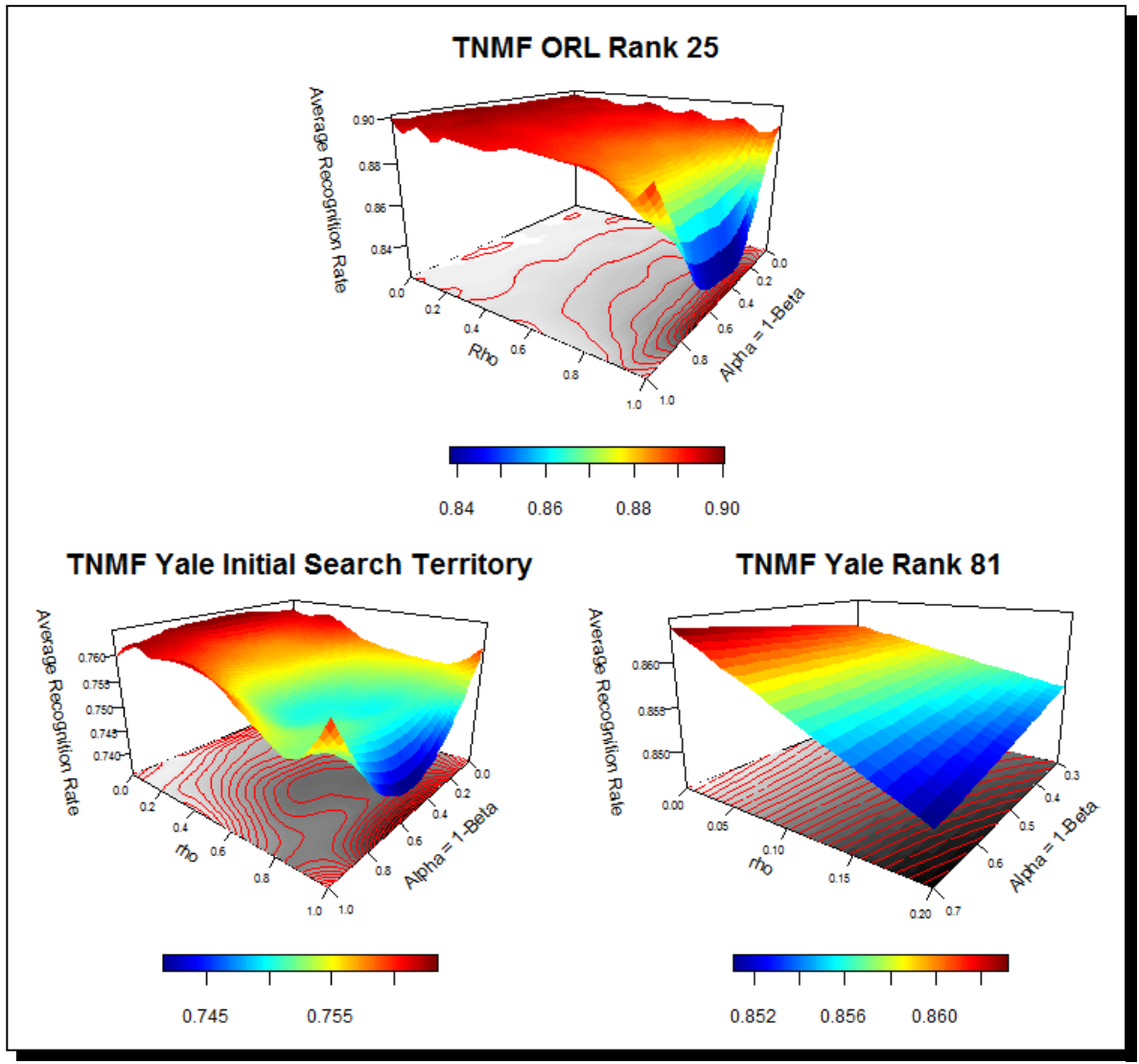

**Figure 5.** (top) The optimal parameter performance surface generated using the TNMF algorithm on the ORL dataset with a factorization rank of $k = 36$; (bottom left) The initial search region for the parameters of the TNMF algorithm using the Yale dataset with $k = 16$; (bottom right) The most promising granular search surface using the TNMF algorithm with $k = 81$ for the Yale dataset

After identifying the optimal parameter settings in each of the aforementioned scenarios, 500 replications were carried out at each setting. The average recognition performances are shown in Figure 6 and parameter settings by algorithm and dataset are provided in Tables 1–3.

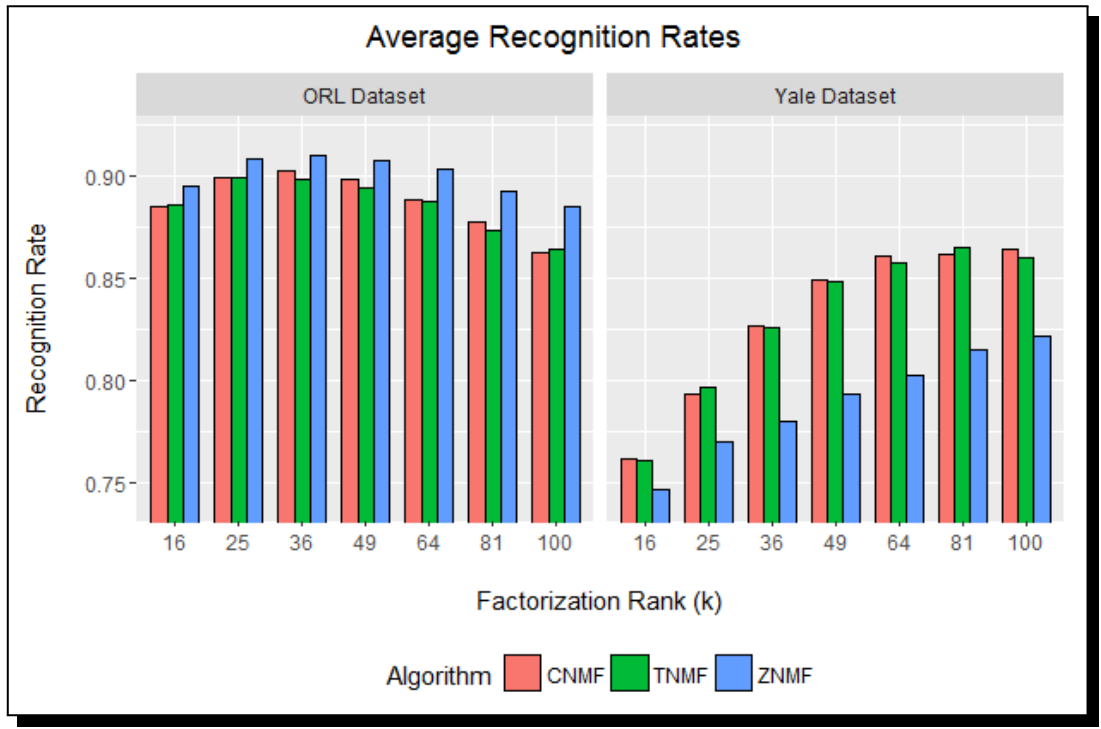

**Figure 6.** Average recognition performance of CNMF, TNMF and ZNMF algorithms using the ORL and Yale datasets

**Table 1.** Zellner Nonnegative Matrix Factorization Optimal Tuning Parameter Settings and Recognition Performances

| Dataset | Rank ($k$) | $g$-prior | $\alpha$ | $\beta$ | Average Recognition |
|---|---|---|---|---|---|
| ORL | 16 | 75.0 | 0.26 | 0.74 | 0.8944 |
| | 25 | 82.0 | 0.40 | 0.60 | 0.9083 |
| | 36 | 83.0 | 0.44 | 0.56 | 0.9097 |
| | 49 | 75.0 | 0.30 | 0.70 | 0.9075 |
| | 64 | 81.0 | 0.59 | 0.41 | 0.9028 |
| | 81 | 75.0 | 0.31 | 0.69 | 0.8924 |
| | 100 | 80.5 | 0.60 | 0.40 | 0.8849 |
| Yale | 16 | 3,500 | 0.30 | 0.70 | 0.7462 |
| | 25 | 3,500 | 0.00 | 1.00 | 0.7694 |
| | 36 | 4,300 | 0.00 | 1.00 | 0.7801 |
| | 49 | 4,500 | 0.30 | 0.70 | 0.7931 |
| | 64 | 3,500 | 0.00 | 1.00 | 0.8022 |
| | 81 | 4,500 | 0.00 | 1.00 | 0.8149 |
| | 100 | 3,500 | 0.00 | 1.00 | 0.8218 |

**Table 2.** Toeplitz Nonnegative Matrix Factorization Optimal Tuning Parameter Settings and Recognition Performances

| Dataset | Rank ($k$) | $\rho$ | $\alpha$ | $\beta$ | Average Recognition |
|---|---|---|---|---|---|
| ORL | 16 | 0.00 | 0.48 | 0.52 | 0.8853 |
|  | 25 | 0.06 | 0.68 | 0.32 | 0.8986 |
|  | 36 | 1.00 | 0.00 | 1.00 | 0.8979 |
|  | 49 | 1.00 | 0.00 | 1.00 | 0.8938 |
|  | 64 | 0.00 | 0.40 | 0.60 | 0.8871 |
|  | 81 | 0.00 | 0.80 | 0.20 | 0.8732 |
|  | 100 | 0.00 | 0.36 | 0.64 | 0.8643 |
| Yale | 16 | 0.15 | 0.70 | 0.30 | 0.7609 |
|  | 25 | 0.00 | 0.56 | 0.44 | 0.7963 |
|  | 36 | 0.00 | 0.70 | 0.30 | 0.8256 |
|  | 49 | 0.00 | 0.30 | 0.70 | 0.8479 |
|  | 64 | 0.00 | 0.42 | 0.58 | 0.8572 |
|  | 81 | 0.00 | 0.64 | 0.36 | 0.8647 |
|  | 100 | 0.00 | 0.70 | 0.30 | 0.8639 |

**Table 3.** Constrained Nonnegative Matrix Factorization Optimal Tuning Parameter Settings and Recognition Performances.

| Dataset | Rank ($k$) | $\alpha$ | $\beta$ | Average Recognition |
|---|---|---|---|---|
| ORL | 16 | 0.44 | 0.56 | 0.8844 |
|  | 25 | 0.32 | 0.68 | 0.8992 |
|  | 36 | 0.24 | 0.76 | 0.9021 |
|  | 49 | 0.44 | 0.56 | 0.8983 |
|  | 64 | 0.73 | 0.27 | 0.8881 |
|  | 81 | 0.44 | 0.56 | 0.8770 |
|  | 100 | 0.20 | 0.80 | 0.8623 |
| Yale | 16 | 0.69 | 0.31 | 0.7616 |
|  | 25 | 1.00 | 0.00 | 0.7935 |
|  | 36 | 0.83 | 0.17 | 0.8262 |
|  | 49 | 1.00 | 0.00 | 0.8491 |
|  | 64 | 1.00 | 0.00 | 0.8602 |
|  | 81 | 1.00 | 0.00 | 0.8617 |
|  | 100 | 1.00 | 0.00 | 0.8636 |

The ZNMF algorithm managed to outperform, on average, both the CNMF and TNMF algorithm across all factorization ranks when used on the ORL dataset. The TNMF and CNMF algorithms remained relatively comparable in terms of their average facial recognition performances across the factorization ranks for the ORL dataset. It should be noted that the ZNMF algorithm achieved the three highest average facial recognition performance on the ORL dataset at relatively small factorization ranks: 36 (90.97%), 25 (90.83%) and 49 (90.75%). This implies that the ZNMF requires less information, in the form of lower factorization ranks, to produce equal and possibly greater recognition performance on the ORL database as other algorithms (Wang *et al*. [19]) produce when provided with more information in the form of higher factorization ranks. This is important for two related reasons: time and storage. A common problem for many nonnegative matrix factorization algorithms in which a large basis matrix is multiplied through with a large weights matrix is the often lengthy time with which it takes a computer to carry out the matrix multiplication. If a matrix factorization algorithm, such as ZNMF, can produce equally accurate results as other matrix factorization algorithms while using relatively lower dimensional basis and weights matrices then we might expect the ZNMF algorithm to take less time in producing its estimation of **X** depending on the complexity of the competing algorithms. Similarly, an algorithm that can produce equally impressive results as another algorithm while using substantially less storage (e.g. lower factorization ranks) may be beneficial.

The pattern in the algorithms' performances was quite different in the Yale dataset than in the ORL dataset. Where ZNMF was clearly superior to the other methods for the ORL dataset, the opposite is true for the Yale dataset. The CNMF and TNMF algorithms drastically outperformed the ZNMF algorithm across all factorization ranks. The CNMF algorithm outperformed the TNMF algorithm at factorization ranks 16, 36, 49, 64 while the TNMF algorithm outperformed the CNMF algorithm at factorization ranks 25, 81 and 100 for the Yale dataset. The highest average facial recognition performance on the Yale dataset (86.47%) occurred when using the TNMF algorithm with a factorization rank of 81. It is interesting that the average recognition rate continues to increase as the factorization rank increases for all three algorithms when used on the Yale dataset. This might imply that the performance of each algorithm might benefit if higher factorization ranks (i.e. > 100) were to be considered in the future. This is quite different from the performances of the algorithms on the ORL dataset which seem to reach their maximums at a factorization rank of 36. This may be due to the fact that, after the resolution reduction, each image in the ORL dataset contains 644 pixel values, while each image in the Yale dataset contains 4880 pixel values after the resolution reduction.

## 6. Conclusion and Future Work

In this paper, we proposed the TNMF algorithm for the assessment of facial recognition and evaluated its capabilities in this regard using both the Cambridge ORL Database of Faces and the Yale Database of Faces across seven different factorization ranks. We compared the facial recognition capabilities of TNMF with a constrained version of the nonnegative matrix algorithm (CNMF) which imposes auxiliary penalties on the solution of (2) and also a second constrained version of nonnegative matrix factorization (ZNMF) which incorporates data-dependent penalties into its auxiliary constraints. We found the ZNMF algorithm outperformed both TNMF and CNMF across all factorization ranks for the ORL dataset and achieved the highest average recognition rate of 90.97% at a factorization rank of 36. We found that TNMF and CNMF compared relatively similarly with respect to their recognition performances when used on the Yale Dataset and also that ZNMF drastically underperformed when compared with the other two algorithms across all factorization ranks for the Yale dataset. We found the TNMF algorithm achieved the highest average recognition rate (86.47%) for the Yale dataset at a factorization rank of 81.

Though our new TNMF algorithm does not impose data-dependent auxiliary constraints, we hope to apply other data-dependent constraints to the nonnegative matrix factorization algorithm. One such possibility would be to use $G_1(\mathbf{W}) = \text{trace}(\mathbf{X}^T\mathbf{W}\mathbf{W}^T\mathbf{X})$ where $\mathbf{X}^T\mathbf{W} \in \mathbb{R}^{n \times k}$, and $G_2(\mathbf{H}) = \text{trace}(\mathbf{X}\mathbf{H}^T\mathbf{H}\mathbf{X}^T)$ where $\mathbf{X}\mathbf{H}^T \in \mathbb{R}^{p \times k}$. Note that $\mathbf{X}^T\mathbf{W}\mathbf{W}^T\mathbf{X}$ is $n \times n$ and is essentially the linear gram matrix of the projected $\mathbf{X}$. $\mathbf{X}\mathbf{H}^T\mathbf{H}\mathbf{X}^T$ is $p \times p$ and resembles the covariance matrix in the input space.

Nonnegative matrix factorization is applicable to a variety of tasks and therefore we also hope to move beyond facial recognition tasks and assess the performance of our collection of algorithms perhaps in text mining, social network analysis or Bioinformatics, specifically for the analysis of microarray data.

## Acknowledgement

Ernest Fokoué wishes to express his heartfelt gratitude and infinite thanks to our lady of perpetual help for her ever-present support and guidance, especially for the uninterrupted flow of inspiration received through her most powerful intercession.

### Competing Interests

The authors declare that they have no competing interests.

### Authors' Contributions

All the authors contributed significantly in writing this article. The authors read and approved the final manuscript.